\documentclass[letterpaper, 10 pt, conference]{ieeeconf}  

\IEEEoverridecommandlockouts                              

\usepackage{amsmath} 
\usepackage[amssymb]{SIunits}  
\usepackage{multirow}
\usepackage{makecell}
\usepackage{booktabs}
\usepackage{graphicx}
\usepackage{caption}
\usepackage{subcaption}
\usepackage{balance}

\newcommand{\fref}[1]{Fig. \ref{#1}}
\newcommand{\tref}[1]{Table \ref{#1}}
\newcommand{\sref}[1]{Section \ref{#1}}

\newcommand{\etal}{\textit{et al. }}
\newcolumntype{C}[1]{>{\centering\arraybackslash}p{#1}}

\title{\LARGE \bf
Towards Real-time Semantic RGB-D SLAM in Dynamic Environments
}

\author{Tete Ji$^{1}$, Chen Wang$^{2}$, and Lihua Xie$^{1}$ 
    \thanks{This work was partly supported by the Delta-NTU Corporate Laboratory for Cyber-Physical Systems funded by the National Research Foundation (NRF) Singapore under its Corporate Laboratory@University Scheme.}
    \thanks{$^{1}$Tete Ji and Lihua Xie are with the School of Electrical and Electronic Engineering, Nanyang Technological University, 50 Nanyang Avenue, Singapore 639798. {\tt\small \{jitete, elhxie\}@ntu.edu.sg}}
    \thanks{$^{2}$Chen Wang is with the Robotics Institute, Carnegie Mellon University, Pittsburgh, PA 15213, USA. {\tt\small chenwang@dr.com}}
}

\begin{document}

\maketitle
\thispagestyle{empty}
\pagestyle{empty}

\begin{abstract}
Most of the existing visual SLAM methods heavily rely on a static world assumption and easily fail in dynamic environments. Some recent works eliminate the influence of dynamic objects by introducing deep learning-based semantic information to SLAM systems. However such methods suffer from high computational cost and cannot handle unknown objects. 
In this paper, we propose a real-time semantic RGB-D SLAM system for dynamic environments that is capable of detecting both known and unknown moving objects. 
To reduce the computational cost, we only perform semantic segmentation on keyframes to remove known dynamic objects, and maintain a static map for robust camera tracking.
Furthermore, we propose an efficient geometry module to detect unknown moving objects by clustering the depth image into a few regions and identifying the dynamic regions via their reprojection errors.
The proposed method is evaluated on public datasets and real-world conditions.
To the best of our knowledge, it is one of the first semantic RGB-D SLAM systems that run in real-time on a low-power embedded platform and provide high localization accuracy in dynamic environments.
\end{abstract}

\section{Introduction}
Simultaneous Localization and Mapping (SLAM) is a fundamental capability for intelligent robotic applications.  It aims to simultaneously estimate the poses of the robot and build a map of the unknown environment from the data of on-board sensors \cite{yuan2021survey, wang2021intensity}. Benefiting from various low-cost and light-weight cameras available in the market, vision-based SLAM, or visual SLAM (vSLAM), has received increasing attention over the past decades. A number of impressive vSLAM systems have been proposed using different type of cameras, such as monocular SLAM \cite{lsdslam, orbslam}, RGB-D SLAM \cite{dvoslam, wang2017non}, and stereo SLAM \cite{stereolsd, stereodso}. Due to direct availability of the depth map and metric scale, RGB-D camera is a popular choice, especially for indoor scenes.

Although the existing vSLAM systems have achieved successful performance in some situations, the majority of these methods heavily rely on a static world assumption, which greatly limits their deployment in real world scenarios. Because dynamic objects such as moving people, animals and vehicles, have negative influence on the pose estimation and map reconstruction. Although robust estimation techniques such as RANSAC can be applied to filter out some of the outliers, the improvement is still limited since they can only handle slightly dynamic scenes and may still fail when the moving objects cover most of the camera view.

\begin{figure}[t]
    \centering
    \includegraphics[width=\linewidth]{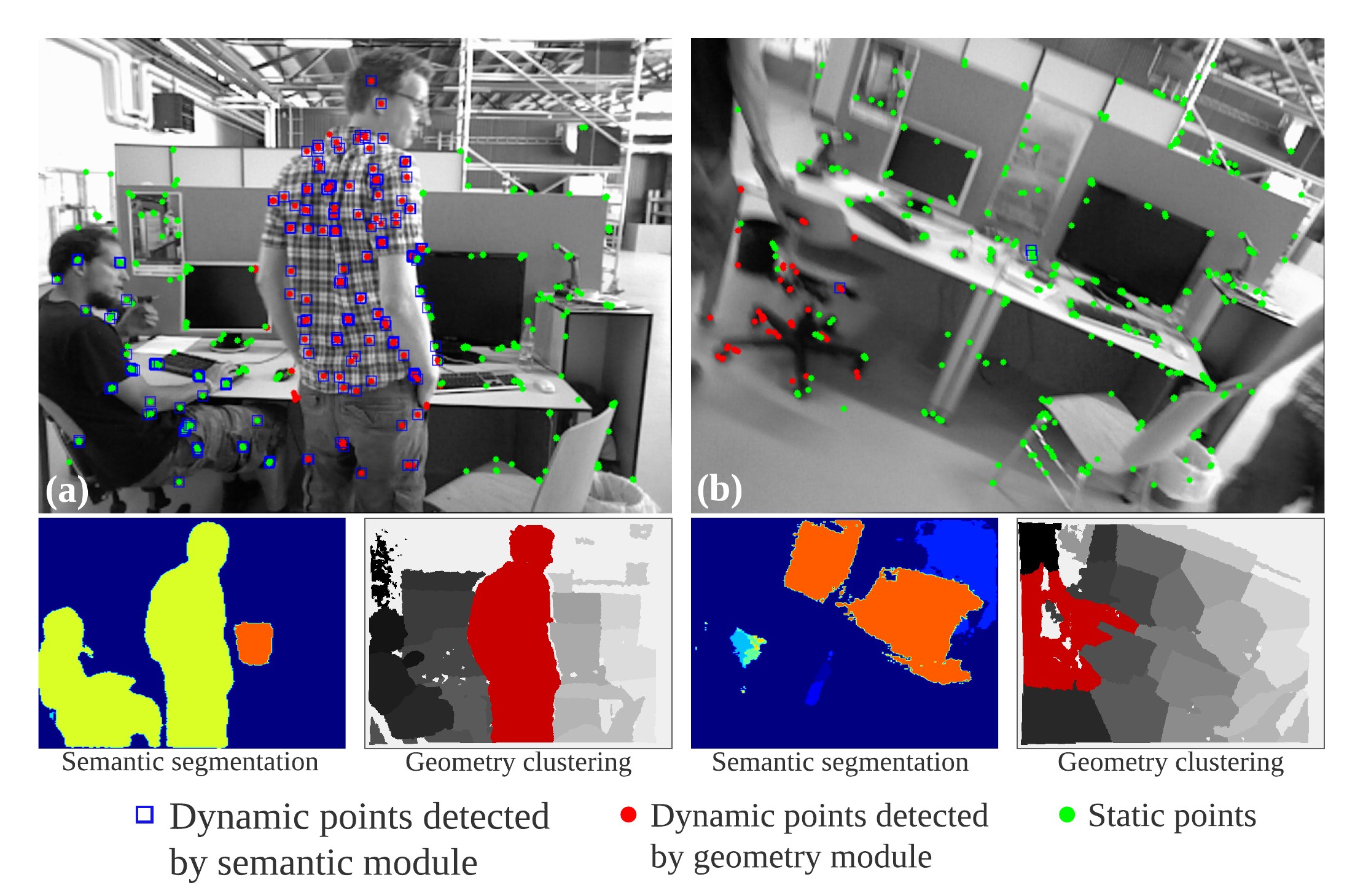}
    \caption{An overview of dynamic objects detection by our proposed method. In case (a), the dynamic feature points associated with both the nearly static sitting person and the walking person are detected by the combination of semantic and geometry information. In case (b), the chair pulled by the person which is an unknown object to the semantic network is also successfully detected by our method.}
    \label{fig:intro}
\end{figure}

Due to recent advances in computer vision and deep learning, semantic information of the environment has been integrated into SLAM systems, such as semantic mapping \cite{semanticfusion, hermans2014dense} and object-level SLAM \cite{fusion++, yang2019cubeslam}. The semantic information is extracted by semantic segmentation which predicts the labels and generates masks for detected objects. By recognizing and removing potentially dynamic objects, the performance of vSLAM in dynamic scenes can be greatly improved \cite{maskfusion, mid-fusion}. However, there still exist two main problems for such methods. One is that most powerful deep neural networks for semantic segmentation such as Mask-RCNN \cite{maskrcnn} are highly computationally expensive and not applicable to real-time and small-scale robotic applications \cite{wang2018correlation}. While for light-weight networks, the segmentation may be less precise and the tracking accuracy will also be affected. The other one is that they can only handle known objects which are labelled in the training set of the network, and may still fail when facing unknown moving objects.

To identify dynamic objects with semantic cues, most of the existing methods perform semantic segmentation on each new frame. This leads to a significant slowdown in camera tracking because the tracking process has to wait until the segmentation completes. 
Therefore, we extract semantic information only from keyframes to remove potentially dynamic objects and maintain a map that contains only static features for camera tracking. Since keyframe and map update can run at low speed in a separate thread, the overall tracking time with semantic segmentation can be significantly reduced. Furthermore, to handle unknown moving objects, we propose an efficient geometry module that does not require prior information about the moving objects. This is done by segmenting the depth image into a few regions using K-Means algorithm and the dynamic regions are identified according to their average reprojection errors. Some examples are demonstrated in \fref{fig:intro}. Different from \cite{staticfusion} which detects dynamic objects using geometry clustering in a dense optimization framework, we identify dynamic regions directly from reprojection errors of sparse features, which leads to faster processing and makes it more robust to dynamic contents.

The main contributions of this paper include:
\begin{itemize}
    \item A keyframe-based semantic RGB-D SLAM system that is capable of reducing the influence of moving objects in dynamic environments. 
    \item An effective and efficient geometry module that deals with unknown moving objects and combines with the semantic SLAM framework.
    \item Extensive evaluations showing that our method provides competitive accuracy compared to the state-of-the-art dynamic SLAM methods while being able to run in real-time on an embedded system.
\end{itemize}

\begin{figure*}[t]
    \centering
    \includegraphics[width=0.9\linewidth]{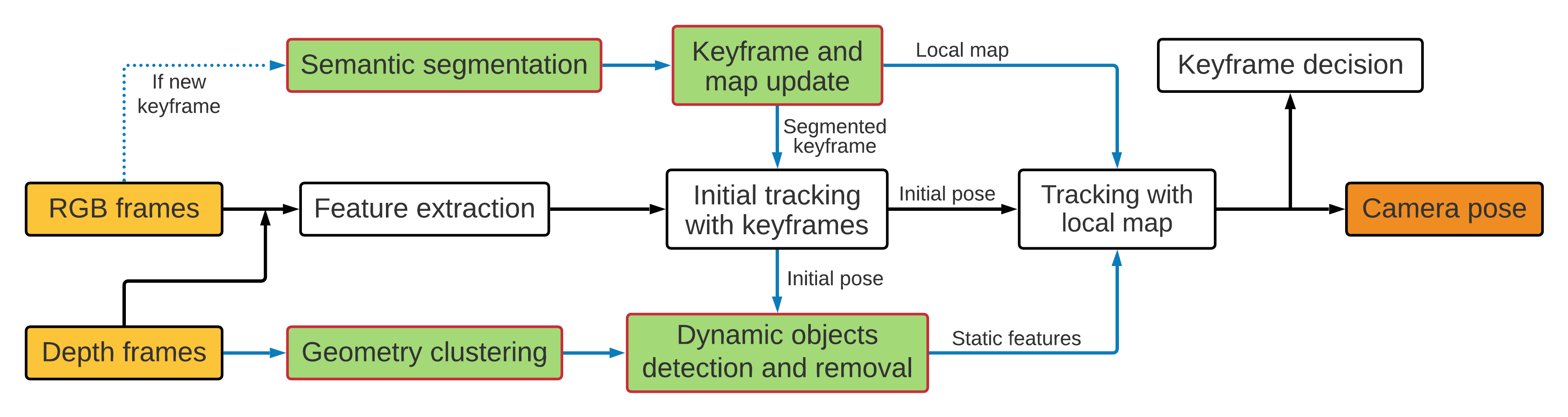}
    \caption{The overall framework of proposed method. The black boxes are the tracking stages modified from ORB-SLAM2. The green-shaded boxes are our added stages to the SLAM system, which include a semantic module and a geometry module. The semantic module performs semantic segmentation on RGB image if it is a new keyframe, and then removes potentially dynamic objects in the keyframe and map. The geometry module performs geometry clustering on depth images and identifies moving objects based on the reprojection errors. After this, only static features are used for pose estimation.}
    \label{fig:framework}
\end{figure*}

\section{Related Work}\label{sec:related work}
To deal with dynamic objects in the environments, a few vSLAM methods have been proposed in recent years. They can be divided into two main directions, one is geometry-based and the other is semantic or learning-based. 

\subsection{Geometry-based Dynamic SLAM}
The main idea of geometry-based methods for dealing with dynamic objects is to treat them as outliers and reject them using robust weighting functions or motion consistency constraints. Sun \etal \cite{motionremoval} propose a motion removal approach that tracks and filters out motion patches in images using particle filter. Kim \etal \cite{bamvo} use a nonparametric model of the background and estimate the ego-motion only based on the estimated background model. StaticFusion \cite{staticfusion} reconstructs the background by estimating a probabilistic segmentation of the image and integrating it in a weighted dense optimization framework.
Zhang \etal \cite{flowfusion} use dense optical flow residuals to detect dynamic regions in the scene and reconstruct the static background using similar framework to StaticFusion. More recently, Dai \etal \cite{point-correlations} propose to use point correlations to distinguish static and dynamic map points based the fact that relative positions between static features are constant over time.

Geometry-based approaches do not need prior information about moving objects and the processing speed is generally faster. They are able to improve the performance of existing SLAM systems to some extent, but still need further efforts. Their accuracy is generally lower than that of semantic-based methods. Moreover, geometry-based methods are unable to build semantic maps of the environment which is useful for advanced robotic applications.

\subsection{Semantic or Learning-based Dynamic SLAM}
Learning-based methods make use of deep neural networks to extract semantic information in the environment by obtaining pixel-wise labels of input images and providing masks for each detected object. The potentially dynamic objects can then be identified from the semantic cues in a single image without the need of multi-frame processing. To get more consistent results, some learning-based methods also utilize conventional geometry information to further distinguish static and dynamic objects among the segmented objects. 
For example, DS-SLAM \cite{ds-slam} combines SegNet and optical flow to identify moving people in the scene and removes dynamic features by checking the moving consistency.
DynaSLAM \cite{dynaslam} combines Mask-RCNN and multi-view geometry to handle both known and unknown moving objects. Although it achieved impressive improvement on public datasets, its computational cost is very high and is mainly for offline operation. In \cite{unifiedframework}, a unified framework is introduced for dynamic SLAM and semantic segmentation which mutually improves the accuracy of each other.

Some methods continuously track the motion of the segmented dynamic objects instead of simply removing them. For example, MID-fusion \cite{mid-fusion} proposes an object-centric approach which incorporates object models in dense tracking formulation to track both object poses and camera motion. Differently, \cite{needforspeed} proposes a novel motion model to track rigid moving objects which does not rely on their 3D models. The motion model and semantic information are then integrated in a factor graph optimization framework for objects and camera tracking.
These methods achieved much improvement in dynamic environments, but all of them are either not able to run in real-time or not able to process unknown moving objects.

\section{Proposed Method}\label{sec:proposed method}
To deal with moving objects in dynamic environments, we make use of both semantic and geometry information from the input RGB-D images. Specifically, we take advantage of the learning-based methods for semantic information to process potentially dynamic objects, and we propose an efficient geometry module to deal with unknown moving objects.
In order to achieve real-time performance, which is vital for robotic applications, we extract semantic cues only from keyframes with a light-weight network. Our system is built on ORB-SLAM2 \cite{orbslam2}, which is a feature-based SLAM system mainly for static environments. As our SLAM is sparse feature based, we only need to reject feature points associated with the moving objects. The overall framework is shown in \fref{fig:framework}, in which the semantic and the geometry module will be presented in the following sections.

\subsection{Semantic Module}
The semantic segmentation is to predict pixel-wise labels and generate masks for detected objects in the input RGB images using deep learning-based methods. To account for both accuracy and computational speed, we adopted the light-weight semantic segmentation network SegNet \cite{segnet} for our semantic module. Using more powerful deep neural networks like Mask-RCNN \cite{maskrcnn} can provide more precise segmentation results, however, the computation will be more expensive. The segmentation network is pre-trained on the PASCAL VOC dataset \cite{pascalvoc}, which contains 20 classes of objects. Among these objects, we only deal with those that are highly movable or potentially dynamic, such as person, car, bicycle, etc. These objects will be removed from the segmented images and the feature points associated with them will not be used for camera tracking and map building. Instead of performing semantic segmentation for each new frame, which is standard in most existing learning-based dynamic SLAM methods, we do this only when a new keyframe is created. This significantly reduces the computational cost for the semantic module and helps us achieve real-time tracking with semantic information. Moreover, this process is performed in a separate thread, and hence does not have much influence on the overall tracking time.

To retain more information for tracking, some learning-based approaches further identify whether the segmented movable objects are moving or not. However for consistent long-term map building this is not necessary as these movable objects are undesired even when they are temporarily static. Hence we leave out this procedure in our semantic module. This shows another advantage of semantic information as it is able to recognize those undesired objects in the scene which are temporarily static and cannot be detected by conventional geometry-based approaches.

\begin{figure*}[t!]
  \centering
  \includegraphics[trim=0 0.4cm 0 0.3cm, clip, width=\linewidth]{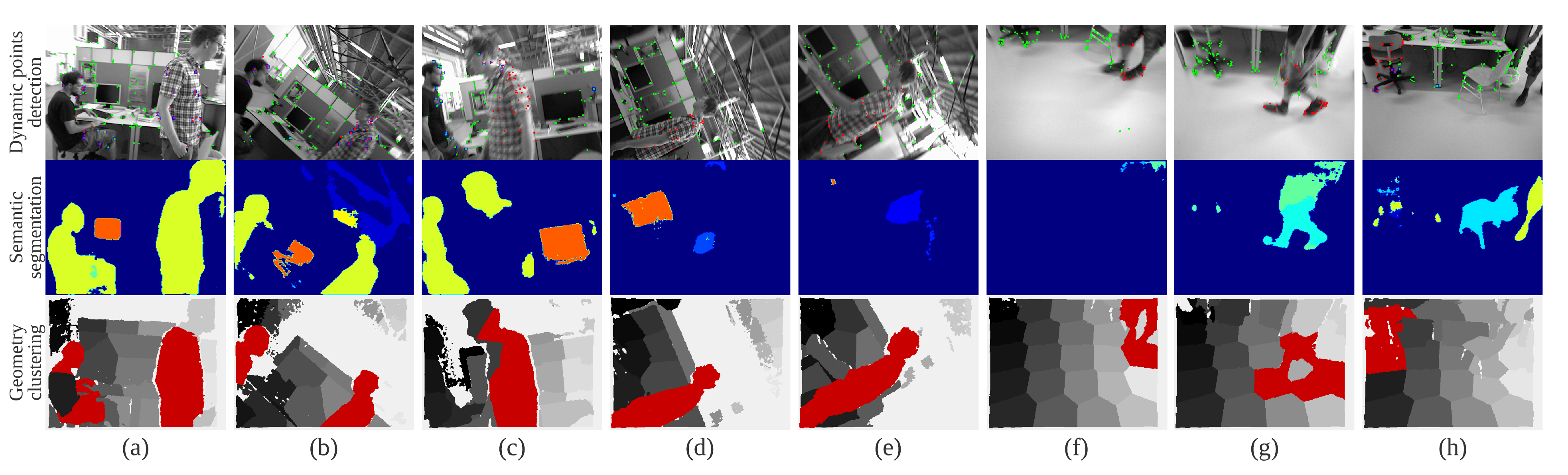}
  \caption{Example results of dynamic objects detection. The first row shows dynamic features detected by the proposed semantic module (points with blue rectangle) and by the geometry module (points in red). The second row is the corresponding semantic segmentation result. The third row shows the geometry clustering results from depth images and the dynamic clusters are highlighted in red. (a) and (b) show that both modules detected dynamic objects. (c)-(h) demonstrate that semantic segmentation failed while geometry module succeeded. Frames are taken from the sequence \textit{fr3/walking\_rpy}.}
  \label{fig:illustration}
\end{figure*}
  

\subsection{Geometry Module} 
Since semantic information alone can only detect a fixed number of object classes that are labelled in the training set, the tracking and mapping will still be affected in the presence of unknown moving objects. Therefore, to handle unknown moving objects, we propose to combine a geometry module that does not require prior information about the objects. 

To achieve this, we first segment each new depth image into $N$ clusters using K-Means algorithm, where points close to each other in 3D space are grouped together. Each cluster is assumed to be a surface of an object and points within a cluster share the same motion constraint. Because a single object could be segmented into a few clusters, the object is not required to be rigid, while most semantic SLAM methods have this rigidity assumption. For each cluster $c_j$, we compute an average reprojection error $r_j$ of all feature points $\mathbf{u}^{'}_i$ inside the cluster against their matched correspondences $\mathbf{P}_i$ in 3D space, as defined in \eqref{eq:error}, where $m$ is the number of matched features in $c_j$,  $\mathbf{T}_{wc}$ is the camera pose, ${\pi}$ represents the camera projection model, and $\rho$ is a penalty function. 

\begin{equation}\label{eq:error}
    r_j=\frac{1}{m}\sum_{i\in{c_j}}^{m} \rho\left(\|\mathbf{u}^{'}_i - {\pi}(\mathbf{T}_{wc}{\mathbf{P}_i})\|^2\right)
\end{equation}

When the error of a cluster is relatively larger than the others, we mark this cluster as dynamic. All feature points in dynamic clusters will be removed and not be involved in camera pose estimation. 
Comparing to identifying the dynamic state for each single feature point, our cluster-wise manner is more effective and efficient. Moreover, it prevents false detection cased by the measurement noise from a single point. It also allows us to approximate a rough shape of the moving objects via the geometry clustering.
Some results of our method can be viewed in the third row of \fref{fig:illustration}, where dynamic clusters are highlighted in red. This module can work independently without semantic information and hence can detect unknown moving objects. 

The geometry module can also be regarded as a backup when semantic segmentation fails in detecting moving objects.
This is because there exist situations that are challenging for semantic segmentation and may cause segmentation errors or failures, especially for light-weight networks. Some examples are presented in \fref{fig:illustration}. 
It can be seen that although both modules can successfully detect moving objects in (a) and (b), the semantic segmentation fails in (c) to (g) due to large rotations, motion blurs, or only part of the objects is present in the images. These situations introduce difficulties to semantic segmentation, while the geometry module can work as normal.
Taking (c) as an example, the semantic segmentation network fails in recognizing a moving person due to severe motion blur caused by the fast relative motion of the person. While the features on this person are successfully detected by our geometry module because the reprojection errors of the dynamic parts are larger than the rest static regions.
Note that we only plot tracked map points in the figure. The dynamic moving objects detected by the geometry module sometimes are not complete because there exist few matched map points in some regions of the objects. This is also a limitation of geometry methods which are difficult to detect a full boundary of moving objects.

An interesting thing we found during experiment is that some semi-dynamic objects could also be identified.
An example is shown in \fref{fig:illustration} (h), where the left chair is identified to be dynamic. The reason is that the chair is currently static but its position was changed when revisiting it. This could be helpful for long-term consistent map building.

\subsection{Keyframe and Local Map Update}
Most of the existing learning-based dynamic SLAM methods perform semantic segmentation for all new frames, which can only run at very low speed or offline mode. Different from that, we extract semantic information only from keyframes. Since a new frame is tracked with keyframes and local map, we only need to ensure that the segmented keyframes and the local map contain only static parts of the scene. The keyframe selection strategy is inherited from the original ORB-SLAM2 system and details will be omitted here.
When a new keyframe is selected during tracking, we perform semantic segmentation in a separate thread and remove dynamic feature points. The local map is also updated by removing the corresponding dynamic map points. In this way, we maintain a keyframe database and a map that contain only static features and map points.

\subsection{Tracking}
We perform a two-stage tracking for each new frame inspired by ORB-SLAM2. We first perform an initial tracking with a recent keyframe that has the largest overlap with the current frame to obtain an initial pose estimation. Since the keyframes have been refined with potentially dynamic objects removed, this initial estimation would be more reliable.
The initial pose estimate is then used in the geometry module for dynamic objects detection.
After removing dynamic points in the current frame by the geometry module, a more accurate pose estimation using local bundle adjustment is obtained by tracking with all the local map points observed in the current frame. As potentially dynamic map points are also removed in the local map by the semantic module, the influence of dynamic objects is further reduced and hence the pose estimation is more robust and accurate.

\section{Experiments and Results}\label{sec:results}

\paragraph{Overview} 
The proposed method is tested on the TUM RGB-D dataset \cite{tumbenchmark} which is widely used for the evaluation of RGB-D SLAM. Since our system is built on ORB-SLAM2, we use it as a baseline to demonstrate our improvement. We first evaluate the effect of different modules introduced in this paper.
We then compare our results with the state-of-the-art dynamic SLAM methods including both geometry-based and learning-based approaches.
The runtime analysis is also presented to show the efficiency of our method.
Furthermore, we demonstrate the performance of our method with a live camera in real scenarios.

\paragraph{Implementation}
For semantic segmentation, we use SegNet \cite{segnet} with its default settings as described in \sref{sec:proposed method}-A. For geometry module, we set the number of clusters $N=24$ at an image resolution of $640\times480$. Since false negatives (dynamic clusters marked as static) are unfavorable compared to false positives (static clusters marked as dynamic), we choose a relatively lower threshold for determining dynamic clusters, which depends on the average reprojection error of matched features. 
All experiments are conducted on NVidia Jetson AGX Xavier with 8-core ARM v8.2 64-bit CPU, 512-core Volta GPU and 16G RAM, which is a low-power and small-size embedded platform designed for intelligent robotic applications. Our algorithm is tested on Ubuntu 18.04 with the Robot Operating System (ROS). To the best of our knowledge, our method is the first semantic RGB-D SLAM system for dynamic environments that is able to run in real-time on such an embedded platform. 

\paragraph{TUM Dataset}
The TUM RGB-D dataset contains 39 sequences captured in indoor environments with an RGB-D camera, and along with ground truth trajectories obtained from a high-accuracy motion capture system. To demonstrate the performance in dynamic environments, we selected a few sequences containing dynamic objects for evaluation.
The \textit{fr3/sitting} (\textit{fr3/s} for short) sequences are slightly dynamic, where two persons sit at a desk with slight body motions. The \textit{fr3/walking} (\textit{fr3/w} for short) sequences are highly dynamic, where two persons walk around a desk. The \textit{walking} sequences are most challenging since the dynamic objects occupy a large portion of the camera view. 
There are four types of camera motions as indicated in the sequence names, remaining almost static (\textit{static}), moving along three directions (\textit{xyz}), rotating along principal axes (\textit{rpy}), and moving along a half sphere with $1$-$\meter$ diameter (\textit{half}). 

\paragraph{Evaluation Metrics} 
The error metrics for evaluation are the commonly-used Root-Mean-Square-Error (RMSE) of the Absolute Trajectory Error (ATE) in $\meter$, and RMSE of the Relative Pose Error (RPE) which comprises the translational drift in $\meter/\second$ and rotational drift in $\degree/\second$. The ATE measures the the global consistency of the trajectory and the RPE measures the odometry drift per second.

\begin{table}[!t]
    \centering
	\caption{Evaluation of ATE on the TUM dataset using the proposed method with different configurations [$\meter$]}
	\begin{tabular}{l|c|c c c}
		\toprule
		Sequence & ORB-SLAM2 & \makecell{Geometry\\module} & \makecell{Semantic\\module} & \makecell{Combined\\system} \\
		\midrule
		fr3/s\_xyz     & \textbf{0.0094} & 0.0100 & 0.0116 & 0.0117 \\
		fr3/s\_half    & 0.0242 & 0.0187 & 0.0174 & \textbf{0.0172} \\ [1ex]
		fr3/w\_static  & 0.1908 & \textbf{0.0111} & 0.0177 & \textbf{0.0111} \\
		fr3/w\_rpy     & 0.8467 & 0.2143 & 0.0373 & \textbf{0.0371} \\
		fr3/w\_xyz     & 0.4662 & 0.0290 & 0.0217 & \textbf{0.0194} \\ 
		fr3/w\_half    & 0.4430 & 0.0362 & 0.0295 & \textbf{0.0290} \\
		\bottomrule
	\end{tabular}
	\label{tab:self-comparison}
\end{table}

\subsection{The Effect of Different Modules}
\begin{figure}[!t]
    \centering
    \begin{subfigure}[b]{0.49\linewidth}
      \includegraphics[trim=0 0.2cm 0 0, clip, width=\linewidth]{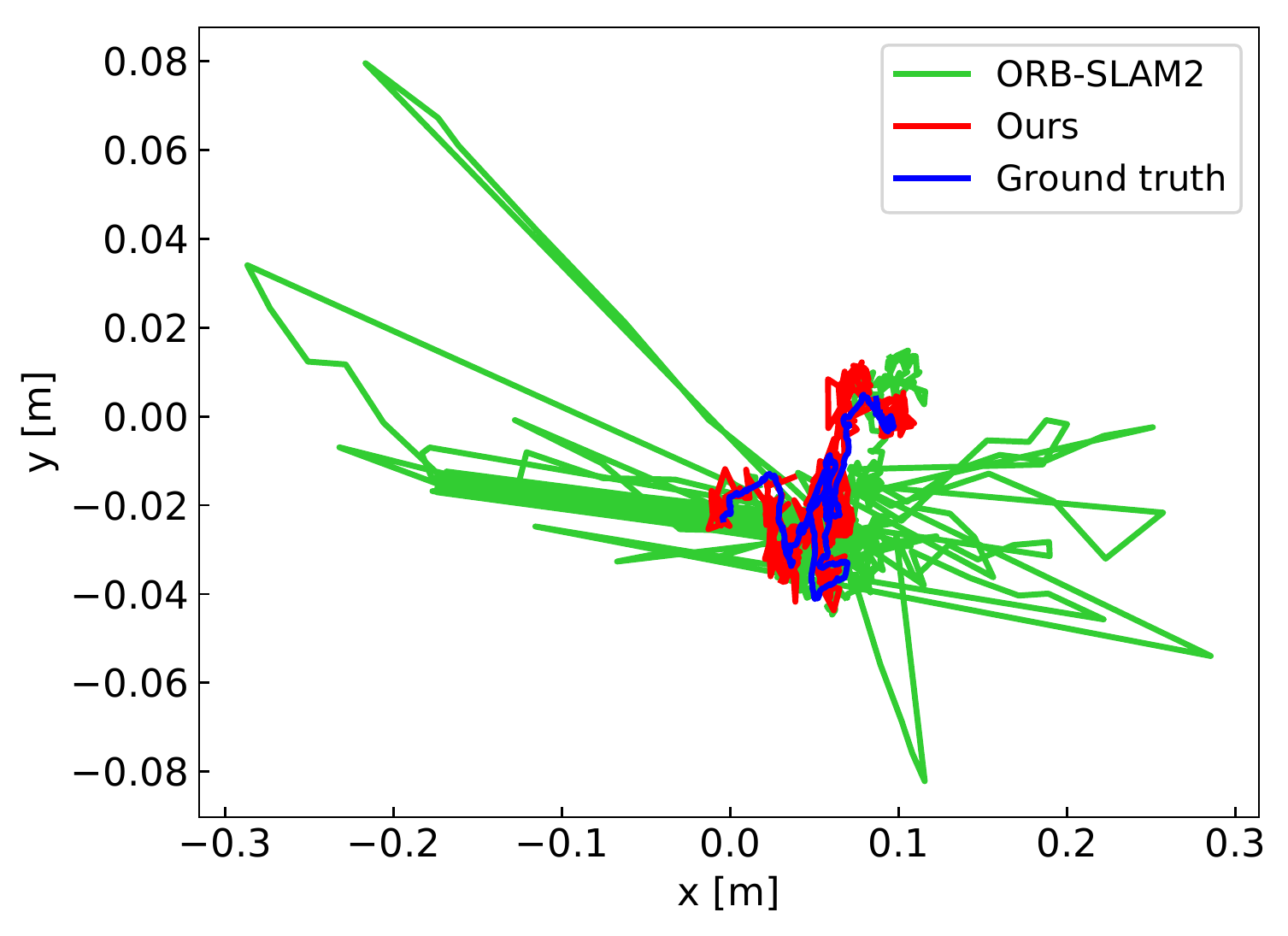}
      \caption{fr3/walking\_static}
    \end{subfigure}
    \begin{subfigure}[b]{0.49\linewidth}
      \includegraphics[trim=0 0.2cm 0 0, clip, width=\linewidth]{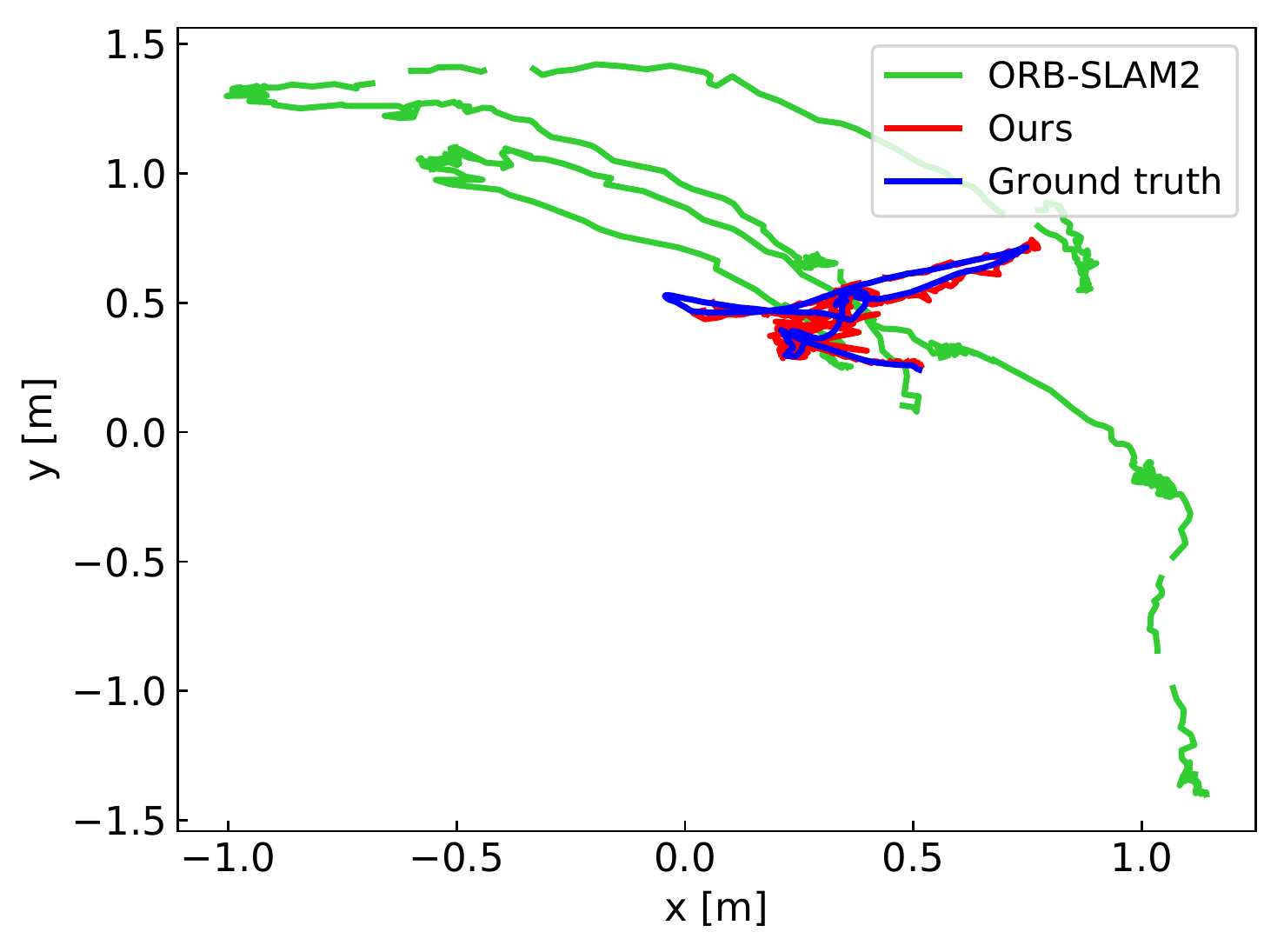}
      \caption{fr3/walking\_rpy}
    \end{subfigure}
    \begin{subfigure}[b]{0.49\linewidth}
      \includegraphics[trim=0 0.2cm 0 0, clip, width=\linewidth]{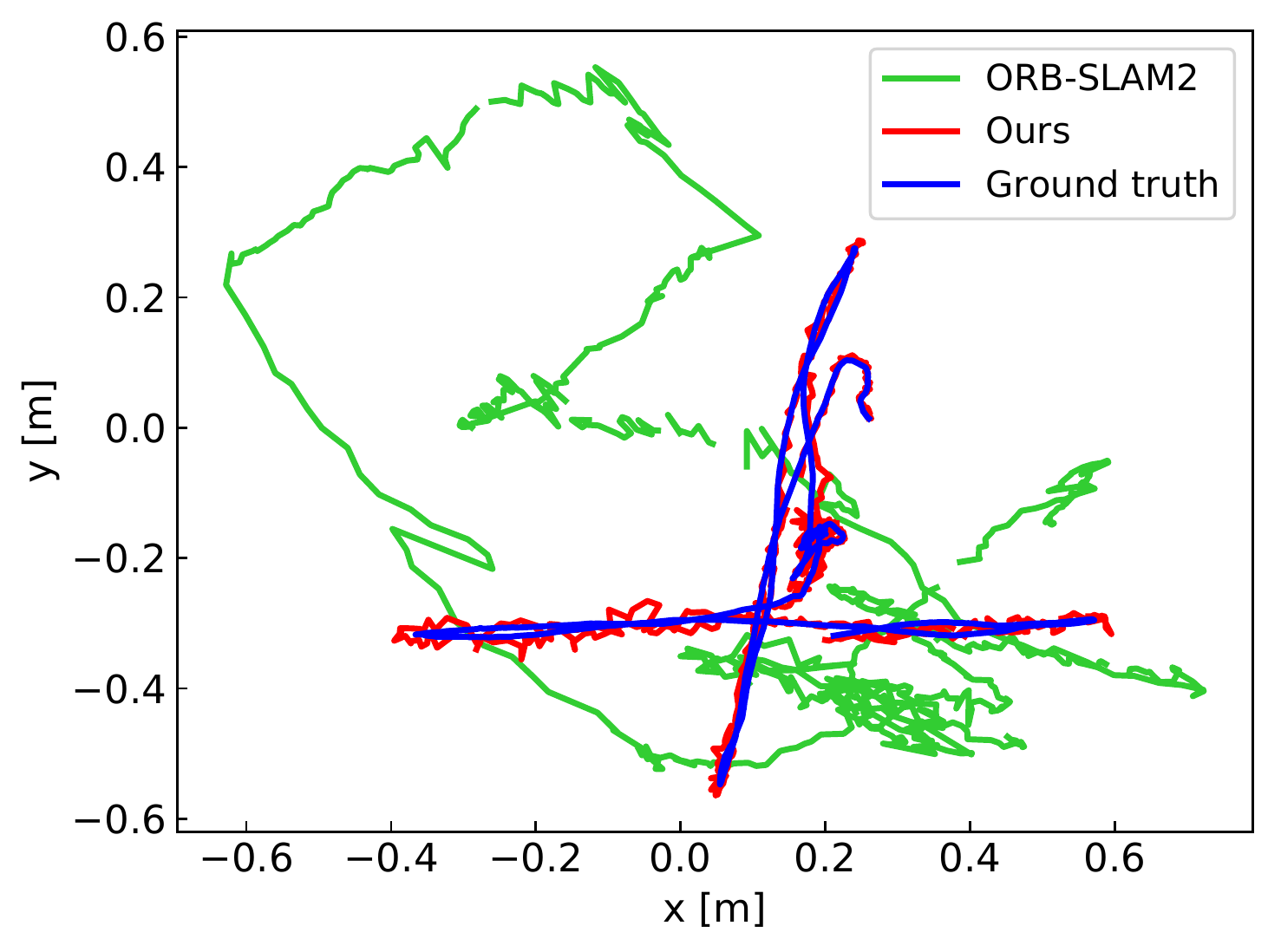}
      \caption{fr3/walking\_xyz}
    \end{subfigure}
    \begin{subfigure}[b]{0.49\linewidth}
      \includegraphics[trim=0 0.2cm 0 0, clip, width=\linewidth]{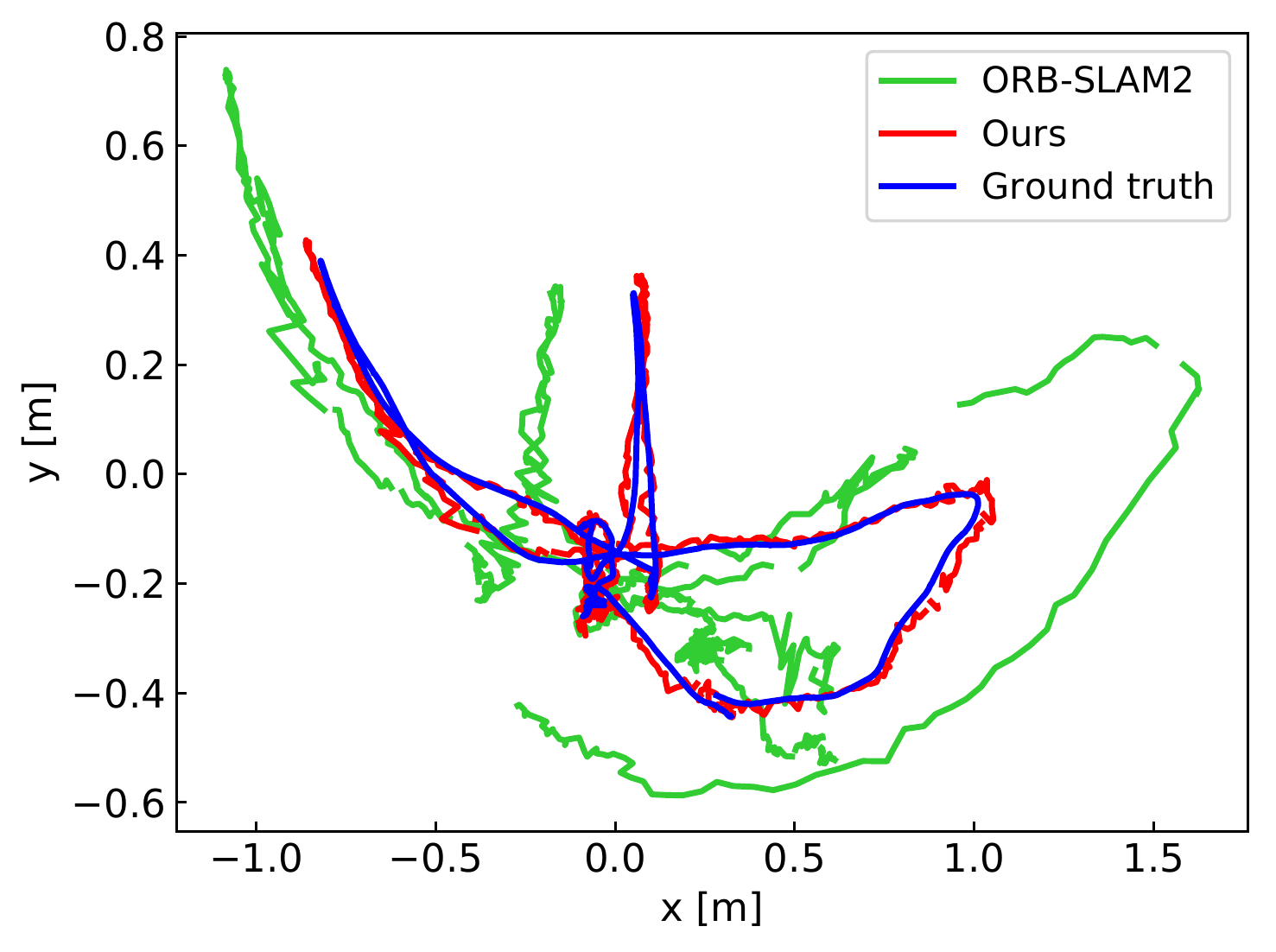}
      \caption{fr3/walking\_half}
    \end{subfigure}
    \caption{Comparison of trajectories estimated by ORB-SLAM2 and the proposed method against ground truth.}
    \label{fig:trajectory}
\end{figure}

We first evaluate the accuracy of our system with different configurations including the geometry module, the semantic module, and their combined framework. The comparison of RMSE of the ATE against the baseline ORB-SLAM2 is shown in Table \ref{tab:self-comparison}. 
For slightly dynamic sequences, our proposed method provides similar results to ORB-SLAM2 since ORB-SLAM2 can handle these situations successfully by the RANSAC algorithm, and hence the improvement margin is limited. While for highly dynamic sequences, both of our semantic and geometry module achieve significant improvement in accuracy, and the proposed combined system achieves the best results. In fact, ORB-SLAM2 failed in tracking most of the time because of the dynamic objects. Therefore by removing the dynamic parts, we can obtain much more accurate camera pose estimations. This can also be seen in the trajectory estimation shown in \fref{fig:trajectory}.

\subsection{Comparison with State-of-the-arts}
We compare our results with state-of-the-art geometry-based dynamic SLAM methods, MR-DVO \cite{motionremoval}, SPW \cite{staticpointweight}, StaticFusion \cite{staticfusion}, DSLAM \cite{point-correlations}, and learning-based methods, MID-Fusion \cite{mid-fusion}, EM-Fusion \cite{emfusion}, DS-SLAM \cite{ds-slam}, and DynaSLAM \cite{dynaslam}. The comparisons of the ATE and RPE are summarized in \tref{tab:ate-comparison} and \tref{tab:rpe-comparison} respectively.
It can be seen that our method provides competitive results in all the dynamic sequences and outperforms all other dynamic SLAM methods except for DynaSLAM, which combines multi-view geometry in a semantic framework. However, it should be noted that DynaSLAM offers offline static map creation and is not able to run in real-time due to its time-consuming Mask-RCNN network and region growing algorithm. On the contrary, our method achieves real-time operation while providing very close results compared to it.

\begin{table*}[!t]
  \centering
  \caption{Comparison of RMSE of the Absolute Trajectory Error (ATE). The best results are highlighted in bold and the second-best are underlined. We use the results published in their original papers when applicable. [$\meter$]}
  \begin{tabular}{l | C{1.5cm} c c c | c c c c | C{1cm}}
    \toprule
    \multirow{2}{*}{Sequence} &\multicolumn{4}{c|}{Geometry-based methods} &\multicolumn{4}{c|}{Learning-based methods} & \multirow{2}{*}{Ours} \\
     & MR-DVO & SPW & StaticFusion & DSLAM & MID-Fusion & EM-Fusion & DS-SLAM & DynaSLAM & \\
    \midrule
    fr3/sitting\_xyz        & 0.0482 & 0.0397 & 0.040 & \textbf{0.0091} &  0.062 & 0.037 & - & 0.015 & \underline{0.0117} \\
    fr3/sitting\_half & 0.0470 & 0.0432 & 0.040 & \underline{0.0235} & 0.031 & 0.032 & - & \textbf{0.017} &  \textbf{0.0172} \\ [1ex]
    fr3/walking\_static     & 0.0656 & 0.0261 & 0.014 & 0.0108 & 0.023 & 0.014 & \underline{0.0081} & \textbf{0.006} & 0.0111 \\
    fr3/walking\_rpy        & 0.1333 & 0.1791 & - & 0.1608 & - & - & 0.4442 & \textbf{0.035} & \underline{0.0371} \\
    fr3/walking\_xyz        & 0.0932 & 0.0601 & 0.127 & 0.0874 & 0.068 & 0.066 & 0.0247 & \textbf{0.015} & \underline{0.0194} \\
    fr3/walking\_half & 0.1252 & 0.0489 & 0.391 & 0.0354 & 0.038 & 0.051 & 0.0303 & \textbf{0.025} & \underline{0.0290} \\
    \bottomrule
  \end{tabular}
  \label{tab:ate-comparison}
\end{table*}

\begin{table*}[!t]
  \centering
  \caption{Comparison of RMSE of the Relative Pose Error (RPE) in translational drift and rotational drift. $^*$ indicates learning-based methods. Only a few learning-based methods reported RPE in their original papers.}
  \begin{tabular}{l | c c c c |c| c c c c |c}
    \toprule
    \multirow{2}{*}{Sequence} &\multicolumn{5}{c|}{Translational RPE (\meter/\second)} &\multicolumn{5}{c}{Rotational RPE (\degree/\second)} \\
     & SPW & StaticFusion & DSLAM & DS-SLAM$^*$ & Ours & SPW & StaticFusion & DSLAM & DS-SLAM$^*$ & Ours \\
    \midrule
    fr3/sitting\_xyz     & 0.0219 & 0.028 & \textbf{0.0134} & - & 0.0166          & 0.8466 & 0.92 & \textbf{0.5792} & - & 0.5968 \\
    fr3/sitting\_half    & 0.0389 & 0.030 & 0.0354 & - & \textbf{0.0259}          & 1.8836 & 2.11 & 0.8699 & - & \textbf{0.7891} \\  [1ex]
    fr3/walking\_static  & 0.0327 & 0.013 & 0.0141 & \textbf{0.0102} & 0.0117     & 0.8085 & 0.38 & 2.7413 & \textbf{0.2690} & 0.2872 \\
    fr3/walking\_rpy     & 0.2252 & - & 0.2299 & 0.1503 & \textbf{0.0471}         & 5.6902 & - & 4.6327 & 3.0042 & \textbf{1.0587} \\
    fr3/walking\_xyz     & 0.0651 & 0.121 & 0.1266 & 0.0333 & \textbf{0.0234}     & 1.6442 & 2.66 & 2.7413 & 0.8266 & \textbf{0.6368} \\
    fr3/walking\_half    & 0.0527 & 0.207 & 0.0517 & \textbf{0.0297} & 0.0423     & 2.4048 & 5.04 & 0.9854 & \textbf{0.8142} & 0.9650 \\
    \bottomrule
  \end{tabular}
  \label{tab:rpe-comparison}
\end{table*}

\subsection{Runtime Analysis}
To evaluate the efficiency of our proposed method, we measured the average computation time of each module and compared with the state-of-the-art learning-based methods DS-SLAM \cite{ds-slam} and DynaSLAM \cite{dynaslam}, together with the baseline ORB-SLAM2. DS-SLAM and DynaSLAM are also built on ORB-SLAM2. The computation time is obtained using the sequence \textit{fr3/walking\_xyz} on the same embedded platform. The results are listed in \tref{tab:run-time}. 
It can be seen that our method is the only semantic RGB-D SLAM method that achieves real-time\footnote{Real-time in this paper means that a robot is able to process images as fast as a human brain, \textit{i.e.}, 100 \milli\second~per frame \cite{potter1969recognition}.} on an embedded platform for dynamic environments.
DS-SLAM uses the same network as ours and only deals with moving people, but it is still less efficient because it performs semantic segmentation for all frames. DynaSLAM has much higher processing time because its geometry part suffers from time-consuming region growing algorithm and sometimes it takes up to a few seconds to process if there are many dynamic points.

\begin{table}[t]
    \centering
	\caption{Comparison of Computation Time [\milli\second]}
	\begin{tabular}{c C{1.9cm} C{1.9cm} C{1.4cm}}
		\toprule
		Methods & Semantic part & Geometry part & Tracking \\
		\midrule
		ORB-SLAM2 & - & - & 71.84 \\
		DS-SLAM & 75.64 & 47.38 & 148.53 \\
		DynaSLAM & 884.24 & 589.72 & 1144.93 \\
		Ours & 72.36 & 30.14 & 75.82 \\
		\bottomrule
	\end{tabular}
	\label{tab:run-time}
\end{table}

\subsection{Robustness Test in Real Environment}
\begin{figure}[t]
  \centering
  \begin{subfigure}[b]{0.22\linewidth}
    \includegraphics[width=\linewidth]{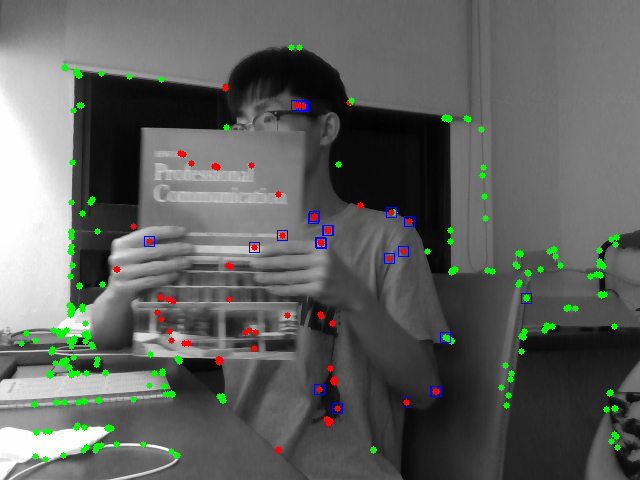}
  \end{subfigure}
  \begin{subfigure}[b]{0.22\linewidth}
    \includegraphics[width=\linewidth]{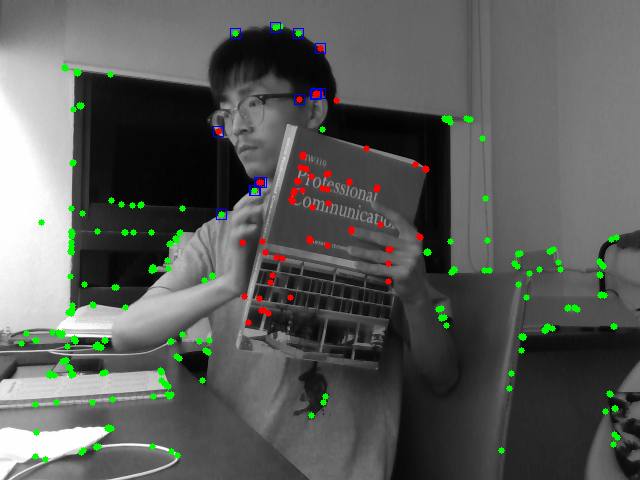}
  \end{subfigure}
  \begin{subfigure}[b]{0.22\linewidth}
    \includegraphics[width=\linewidth]{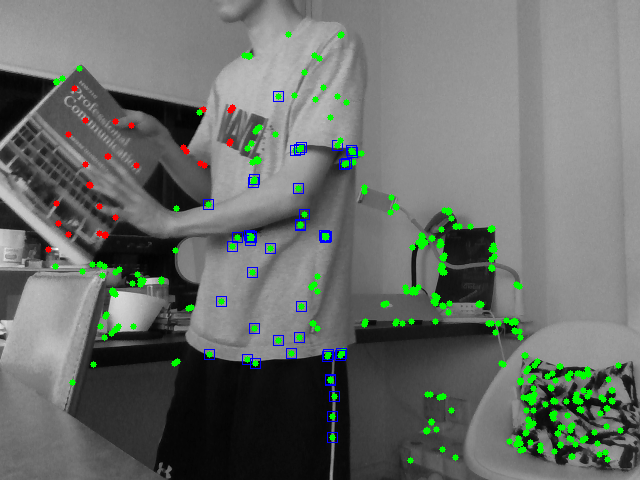}
  \end{subfigure}
  \begin{subfigure}[b]{0.22\linewidth}
    \includegraphics[width=\linewidth]{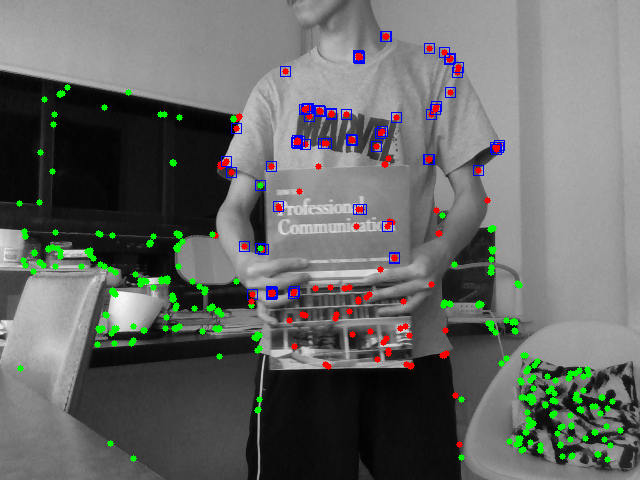}
  \end{subfigure}
  \begin{subfigure}[b]{0.22\linewidth}
    \includegraphics[width=\linewidth]{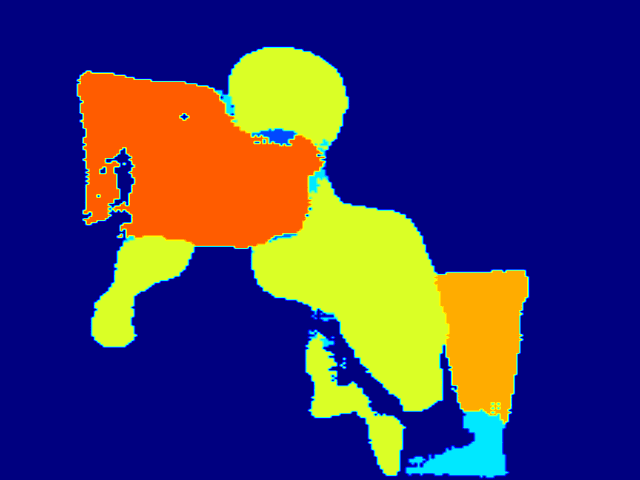}
  \end{subfigure}
  \begin{subfigure}[b]{0.22\linewidth}
    \includegraphics[width=\linewidth]{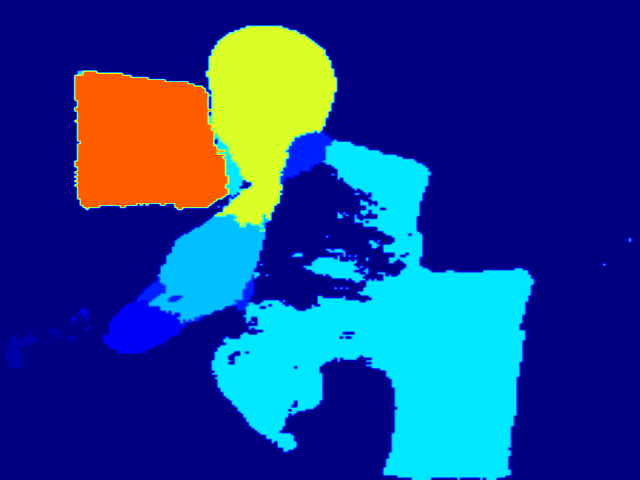}
  \end{subfigure}
  \begin{subfigure}[b]{0.22\linewidth}
    \includegraphics[width=\linewidth]{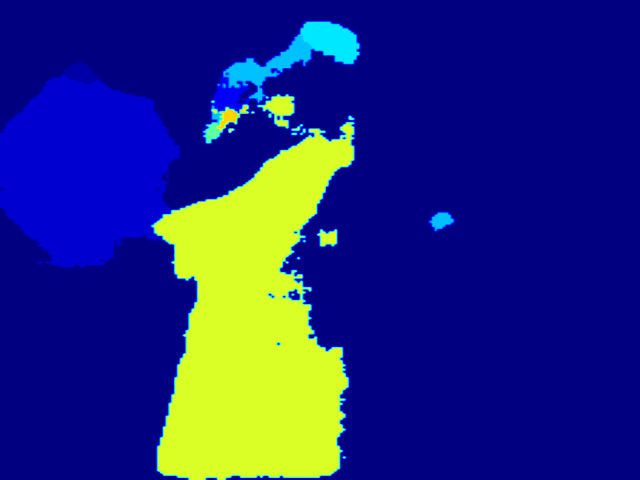}
  \end{subfigure}
  \begin{subfigure}[b]{0.22\linewidth}
    \includegraphics[width=\linewidth]{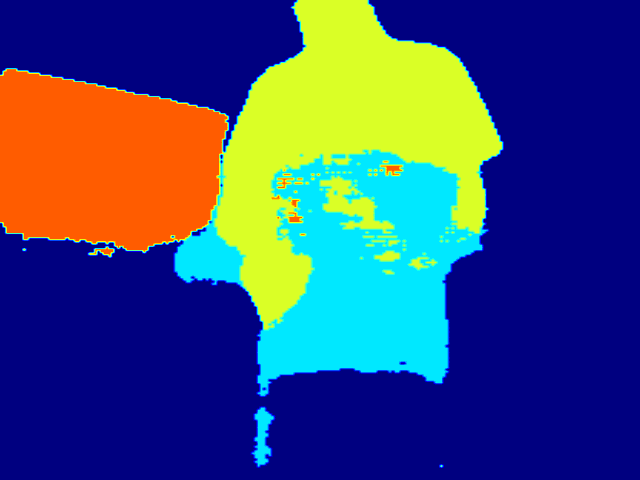}
  \end{subfigure}
  \begin{subfigure}[b]{0.22\linewidth}
    \includegraphics[width=\linewidth]{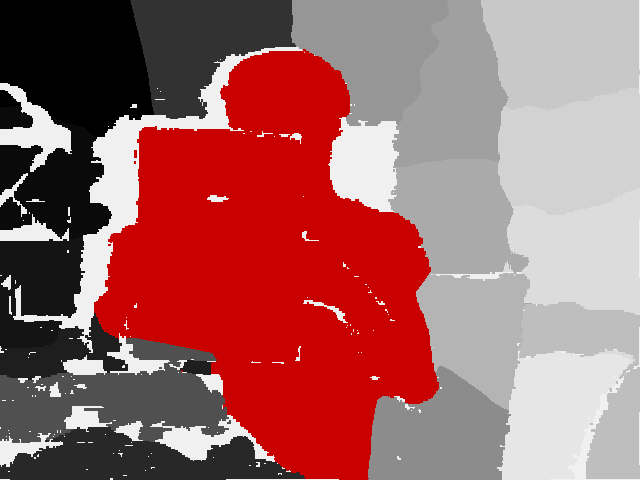}
  \end{subfigure}
  \begin{subfigure}[b]{0.22\linewidth}
    \includegraphics[width=\linewidth]{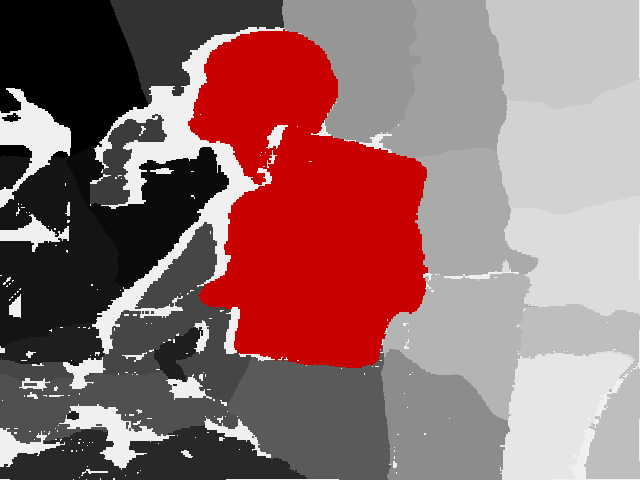}
  \end{subfigure}
  \begin{subfigure}[b]{0.22\linewidth}
    \includegraphics[width=\linewidth]{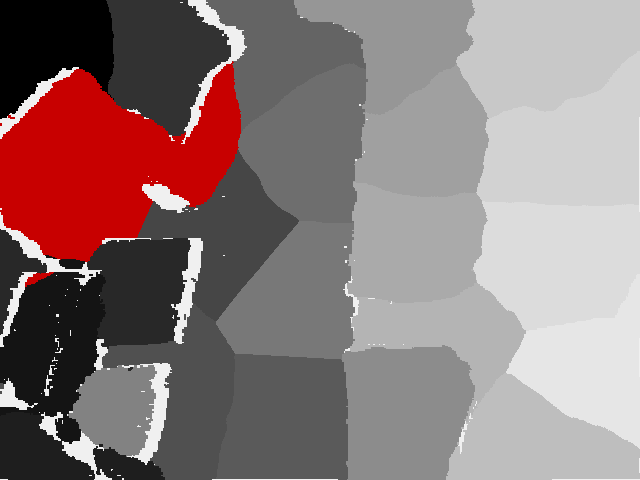}
  \end{subfigure}
  \begin{subfigure}[b]{0.22\linewidth}
    \includegraphics[width=\linewidth]{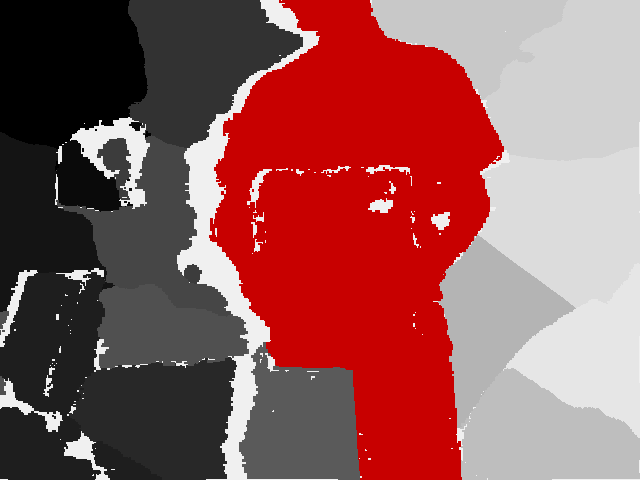}
  \end{subfigure}
  \caption{Experiments with a live RGB-D camera in real dynamic scenarios. The dynamic features associated with both the known object (person) and unknown moving object (book) are successfully detected using our method. }
  \label{fig:real-test}
\end{figure}

We further tested our proposed method with a live RGB-D camera from Intel RealSense to evaluate its robustness in real dynamic environments. In our experiments, a person holding a book is sitting and walking in front of the camera and the camera is holding nearly static. Several screenshots of dynamic points detection results during the real-time test are shown in \fref{fig:real-test}, where the second and third row are segmentation results from the semantic module and our proposed geometry module, respectively. Note that the book is not a labelled object in the network model and hence it cannot be recognized, or sometimes falsely recognized by the semantic module as shown in the second row.
As a compensation process, the geometry module is able to correctly extract the book as a moving object in the test as shown in the third row.
This demonstrates that both the semantic module and geometry module are necessary for robust semantic RGB-D SLAM system in dynamic environments.
The average trajectory estimation error by our method is about $0.012\meter$, while the error by ORB-SLAM2 is about $0.147\meter$ due to large fluctuations caused by the moving objects.
This further demonstrates the effectiveness of our proposed method in the presence of both known and unknown moving objects.


\section{Conclusion}\label{sec:conclusion}
In this paper, we proposed a real-time semantic RGB-D SLAM framework for dynamic environments which is capable of processing both known and unknown moving objects. A keyframe-based semantic module is proposed in order to reduce the computational cost, and an effective geometry module using geometry clustering is introduced to deal with unknown moving objects.
Extensive evaluations demonstrate that our system provides state-of-the-art localization accuracy while still being able to run in real-time on an embedded platform. 
In the future, we plan to build a long-term semantic map of the environment that contains only static parts, which is useful for high level robotic tasks.


\balance
\bibliographystyle{IEEEtran}
\bibliography{IEEEabrv, References}

\end{document}